\documentclass{article}

\usepackage{microtype}
\usepackage{graphicx}
\usepackage{amssymb}
\usepackage{subfigure}
\usepackage{amsmath}
\usepackage{booktabs} 
\usepackage{hyperref}       
\usepackage[usenames, dvipsnames]{color}
\usepackage{graphicx}

\usepackage{macros}

\usepackage{hyperref}

\usepackage[accepted]{icml2018}

\icmltitlerunning{Conditional Neural Processes}

\begin{document}

\twocolumn[
\icmltitle{Conditional Neural Processes}



\icmlsetsymbol{equal}{*}

\begin{icmlauthorlist}
\icmlauthor{Marta Garnelo}{goo}
\icmlauthor{Dan Rosenbaum}{goo}
\icmlauthor{Chris J. Maddison}{goo}
\icmlauthor{Tiago Ramalho}{goo}
\icmlauthor{David Saxton}{goo}
\icmlauthor{Murray Shanahan}{goo,imp}
\icmlauthor{Yee Whye Teh}{goo}
\icmlauthor{Danilo J. Rezende}{goo}
\icmlauthor{S. M. Ali Eslami}{goo}
\end{icmlauthorlist}

\icmlaffiliation{goo}{DeepMind, London, UK}
\icmlaffiliation{imp}{Imperial College London, London, UK}

\icmlcorrespondingauthor{Marta Garnelo}{garnelo@google.com}

\icmlkeywords{Machine Learning, ICML}

\vskip 0.3in
]



\printAffiliationsAndNotice{} 

\begin{abstract}
Deep neural networks excel at function approximation, yet they are typically trained from scratch for each new function. On the other hand, Bayesian methods, such as \acp{gp}, exploit prior knowledge to quickly infer the shape of a new function at test time. Yet \acp{gp} are computationally expensive, and it can be hard to design appropriate priors. In this paper we propose a family of neural models, \acp{dnp}, that combine the benefits of both. \acp{dnp} are inspired by the flexibility of stochastic processes such as \acp{gp}, but are structured as neural networks and trained via gradient descent. \acp{dnp} make accurate predictions after observing only a handful of training data points, yet scale to complex functions and large datasets. We demonstrate the performance and versatility of the approach on a range of canonical machine learning tasks, including regression, classification and image completion.
\end{abstract}

\section{Introduction}
\label{sec:intro}

Deep neural networks have enjoyed remarkable success in recent years, but they require large datasets for effective training~\citep{lake2017building, garnelo2016towards}. One way to mitigate this data efficiency problem is to approach learning in two phases. The first phase learns the statistics of a generic domain, drawing on a large training set, but without committing to a specific learning task within that domain. The second phase learns a function for a specific task, but does so using only a small number of data points by exploiting the domain-wide statistics already learned. Meta-learning with neural networks is one example of this approach~\cite{wang2016learning, reed2017few}.

\begin{figure}[t]
\begin{center}
\centerline{\includegraphics[width=0.85\columnwidth]{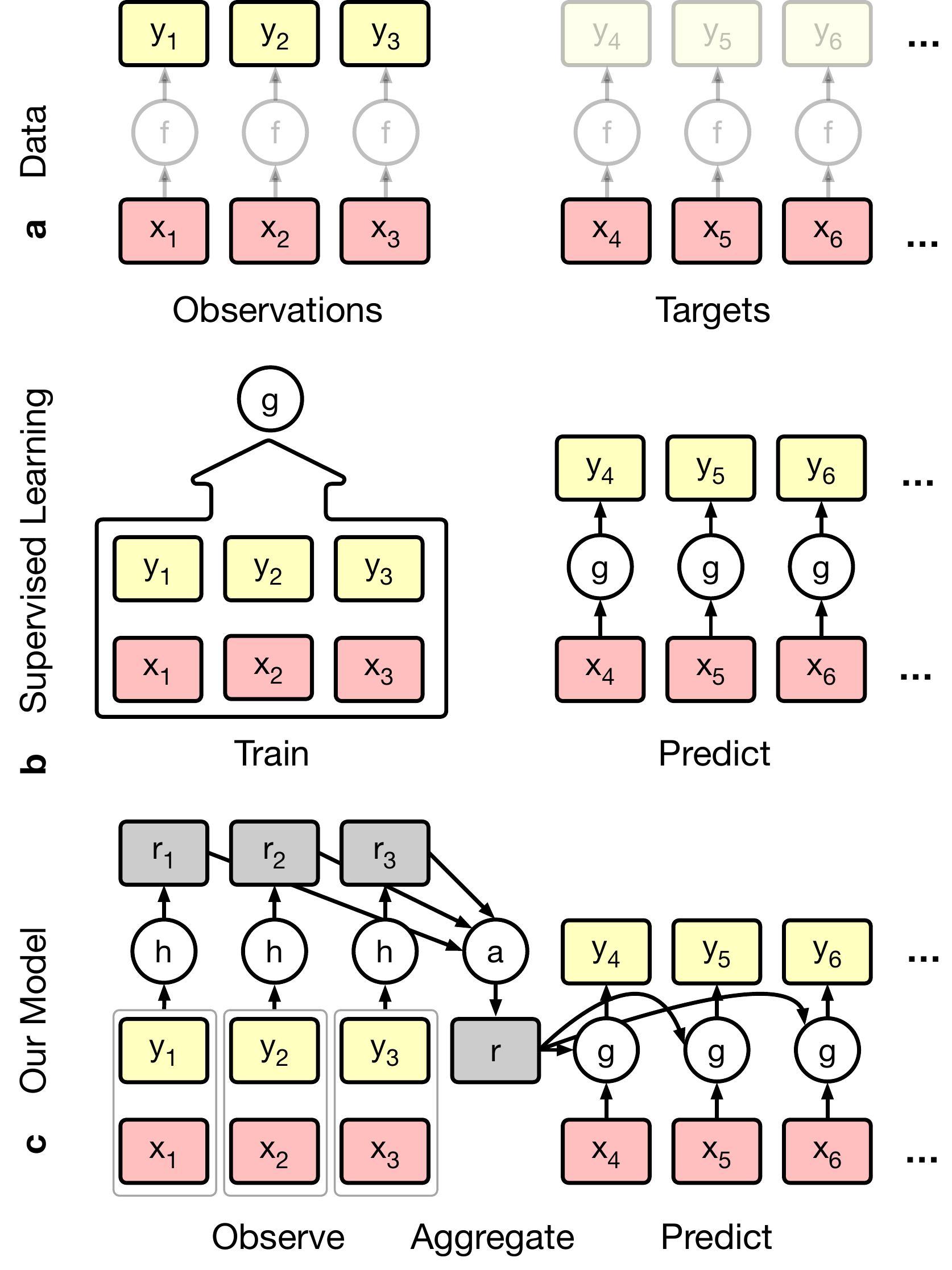}}
\caption{\textbf{\acl{dnp}}. a) Data description b)~Training regime of conventional supervised deep learning models c) Our model.}
\label{fig:model}
\end{center}
\vspace{-2\baselineskip}
\end{figure}

For example, consider supervised learning problems. Many of these can be framed as function approximation given a finite set of observations. Consider a dataset $\{(x_i, y_i)\}_{i=0}^{n-1}$ of $n$ inputs $x_i \in X$ and outputs $y_i \in Y$. Assume that these represent evaluations $y_i = f(x_i)$ of some unknown function $f : X \to Y$, which may be fixed or a realization of some random function. A supervised learning algorithm returns an approximating function $g : X \to Y$ or a distribution over such functions. The aim is to minimize a loss between $f$ and $g$ on the entire space~$X$, but in practice the routine is evaluated on a finite set of observations that are held-out (making them effectively unlabelled). We call these unlabelled data points targets (see figure \ref{fig:model}). Classification, regression, dynamics modeling, and image generation can all be cast in this framework. 

One approach to supervised problems is to randomly initialize a parametric function $g$ anew for each new task and spend the bulk of computation on a costly fitting phase. Prior information that a practitioner may have about $f$ is specified via the architecture of $g$, the loss function, or the training details. This approach encompasses most of deep supervised learning. Since the extent of prior knowledge that can be expressed in this way is relatively limited, and learning cannot be shared between different tasks, the amount of training required is large, and deep learning methods tend to fail when training data is not plentiful.


Another approach is to take a probabilistic stance and specify a  distribution over functions, known as stochastic processes; \aclp{gp} (\acp{gp}) are an example \cite{rasmussen2004gaussian}. On this view, a practitioner's prior knowledge about $f$ is captured in the distributional assumptions about the prior process and learning corresponds to Bayesian inference over the functional space conditioned on the observed values. In the \ac{gp} example, assumptions on the smoothness of $f$ are captured a priori via a parametric kernel function, and $g$ is taken to be a random function distributed according to the predictive posterior distribution. Unfortunately, such Bayesian approaches quickly become computationally intractable as the dataset or dimensionality grows \cite{snelson2006sparse}.




In this work we propose a family of models that represent solutions to the supervised problem, and an end-to-end training approach to learning them, that combine neural networks with features reminiscent of Gaussian Processes.
We call this family of models \emph{\aclp{dnp}}~(CNPs), as an allusion to the fact that they define conditional distributions over functions given a set of observations. The dependence of a \ac{dnp} on the observations is parametrized by a neural network that is invariant under permutations of its inputs. We focus on architectures that scale as $\mathcal{O}(n+m)$ at test time, where $n,m$ are the number of observations and targets, respectively. In its most basic form a \ac{dnp} embeds each observation, aggregates these embeddings into a further embedding of fixed dimension with a symmetric aggregator, and conditions the function $g$ on the aggregate embedding; see Figure \ref{fig:model} for a schematic representation. \acp{dnp} are trained by sampling a random dataset and following a gradient step to maximize the conditional likelihood of a random subset of targets given a random observation set. This encourages \acp{dnp} to perform well across a variety of settings, i.e. $n \ll m$ or $n \gg m$.

This paper is structured as follows. In Section \ref{sec:model} we introduce \aclp{dnp}, propose our implementation, and describe how they can be trained efficiently using standard deep learning libraries. In Section \ref{sec:related} we discuss related work, particularly Bayesian and meta-learning approaches. In Section \ref{sec:experiments} we apply \acp{dnp} to several domains, including regression, classification and image completion, comparing its benefits to classical approaches. We emphasize that although CNPs share some similarities with Bayesian approaches, they do not implement Bayesian inference directly and it is not necessarily true that the conditional distributions will be consistent with respect to some prior process. However, the ability to extract prior knowledge directly from training data tied with scalability at test time can be of equal or greater importance.

\section{Model}
\label{sec:model}

\subsection{Stochastic Processes}

Consider a set $O =\{(x_i, y_i) \}_{i=0}^{n-1} \subset X \times Y$ of pairs of inputs $x_i \in X$ and outputs $y_i \in Y$ and another set $T=\{x_i\}_{i=n}^{n+m-1} \subset X$ of unlabelled points. We call these the set of observations and targets, respectively. We assume that the outputs are a realization of the following process; let $P$ be a probability distribution over functions $f : X \to Y$, formally known as a stochastic process\footnote{$P$ is a measure traditionally defined as the extension of a set of consistent measures on finite collections of $f(x_i)$ for $x_i \in X$. We overload $P$ for the marginal and conditional densities of these measures.}, then for $f \sim P$, set $y_i = f(x_i)$. $P$ defines a joint distribution over the random variables $\{f(x_i)\}_{i=0}^{n + m -1}$, and therefore a conditional distribution $P(f(T) \given O, T)$; our task is to predict the output values $f(x)$ for every $x \in T$ given $O$.

As a motivating example, consider a random \mbox{1-dimensional} function $f \sim P$ defined on the real line (i.e.,~$X \mathrel{\mathop:}= \mathbb{R}$, $Y \mathrel{\mathop:}= \mathbb{R}$). $O$ would constitute $n$ observations of $f$'s value $y_i$ at different locations $x_i$ on the real line. Given these observations, we are interested in predicting $f$'s value at new locations on the real line. A classic assumption to make on $P$ is that all finite sets of function evaluations of $f$ are jointly Gaussian distributed. This class of random functions are known as \acfp{gp}. In this case, the predictive distribution $P(f(T) \given O, T)$ has a simple analytic form defined by prior assumptions on the pairwise correlation structure (specified via a kernel function). The framework of stochastic processes is appealing, because Bayes’ rule allows one to reason consistently about the predictive distribution over $f$ imposed by observing $O$ under a set of probabilistic assumptions. This allows the model to be data efficient, an uncommon characteristic in most deep learning models. However, in practice, it is difficult to design appropriate priors and most interesting examples of stochastic processes are computationally expensive, scaling poorly with $n$ and $m$. This includes GPs which scale as $\mathcal{O}((n+m)^3)$.

\subsection{\acfp{dnp}}

As an alternative we propose \acfp{dnp}, models that directly parametrize conditional stochastic processes without imposing consistency with respect to some prior process. \acp{dnp} parametrize distributions over $f(T)$ given a distributed representation of $O$ of \emph{fixed} dimensionality. By doing so we give up the mathematical guarantees associated with stochastic processes, trading this off for functional flexibility and scalability.

Specifically, given a set of observations $O$, a \ac{dnp} is a conditional stochastic process $Q_{\theta}$ that defines distributions over $f(x)$ for inputs $x \in T$. $\theta$ is the real vector of all parameters defining $Q$. Inheriting from the properties of stochastic processes, we assume that $Q_{\theta}$ is invariant to permutations of $O$ and $T$. If $O^{\prime}, T^{\prime}$ are permutations of $O$ and $T$, respectively, then $Q_{\theta}(f(T) \given O,T) = Q_{\theta}(f(T^{\prime}) \given O, T^{\prime}) = Q_{\theta}(f(T) \given O^{\prime}, T)$. In this work, we generally enforce permutation invariance with respect to $T$ by assuming a factored structure. Specifically, we consider $Q_{\theta}$s that factor $Q_{\theta}(f(T) \given O, T) = \prod_{x \in T} Q_{\theta}(f(x) \given O, x)$. In the absence of assumptions on output space $Y$, this is the easiest way to ensure a valid stochastic process. Still, this framework can be extended to non-factored distributions, we consider such a model in the experimental section. 

The defining characteristic of a \ac{dnp} is that it conditions on $O$ via an embedding of fixed dimensionality. In more detail, we use the following architecture,
\begin{align}
    r_i &= h_{\theta}(x_i,y_i) \qquad \forall (x_i, y_i) \in O\\
    r &= r_1 \oplus r_2 \oplus \ldots r_{n-1} \oplus r_n\\
    \phi_i &= g_{\theta}(x_i, r) \qquad \forall (x_i) \in T
\end{align}
where $h_{\theta} : X \times Y \to \mathbb{R}^d$ and $g_{\theta} : X \times \mathbb{R}^d \to \mathbb{R}^e$ are neural networks, $\oplus$ is a commutative operation that takes elements in $\mathbb{R}^d$ and maps them into a single element of $\mathbb{R}^d$, and $\phi_i$ are parameters for $Q_{\theta}(f(x_i) \given O, x_i) = Q(f(x_i) \given \phi_i)$.
 Depending on the task the model learns to parametrize a different output distribution. This architecture ensures permutation invariance and $\mathcal{O}(n+m)$ scaling for conditional prediction. We note that, since $ r_1 \oplus \ldots \oplus r_n$ can be computed in $\mathcal{O}(1)$ from $ r_1 \oplus \ldots \oplus r_{n-1}$, this architecture supports streaming observations with minimal overhead.
 
 For regression tasks we use $\phi_i$ to parametrize the mean and variance $\phi_i = (\mu_i, \sigma_i^2)$ of a Gaussian distribution $\mathcal{N}(\mu_i, \sigma_i^2)$ for every $x_i \in T$. For classification tasks  $\phi_i$ parametrizes the logits of the class probabilities $p_c$ over the $c$ classes of a categorical distribution. In most of our experiments we take $a_1 \oplus \ldots \oplus a_n$ to be the mean operation $(a_1 + \ldots + a_n)/n$.

\subsection{Training \acp{dnp}}

We train $Q_{\theta}$ by asking it to predict $O$ conditioned on a randomly chosen subset of $O$. This gives the model a signal of the uncertainty over the space $X$ inherent in the distribution $P$ given a set of observations. More precisely, let $f \sim P$, $O = \{(x_i, y_i)\}_{i=0}^{n-1}$ be a set of observations, $N \sim \mathrm{uniform}[0, \ldots, n-1]$. We condition on the subset $O_N = \{(x_i, y_i)\}_{i=0}^{N} \subset O$, the first $N$ elements of $O$. We minimize the negative conditional log probability
\begin{equation}
\begin{aligned}
\mathcal{L}(\theta) = - \mathbb{E}_{f \sim P} \Big[ \mathbb{E}_{N}\Big[ \log Q_{\theta}(\{y_i\}_{i=0}^{n-1} | O_N, \{x_i\}_{i=0}^{n-1})\Big] \Big]
\end{aligned}
\end{equation}
Thus, the targets it scores $Q_{\theta}$ on include \emph{both} the observed and unobserved values. In practice, we take Monte Carlo estimates of the gradient of this loss by sampling $f$ and $N$. 

This approach shifts the burden of imposing prior knowledge from an analytic prior to empirical data. This has the advantage of liberating a practitioner from having to specify an analytic form for the prior, which is ultimately intended to summarize their empirical experience. Still, we emphasize that the $Q_{\theta}$ are not necessarily a consistent set of conditionals for all observation sets, and the training routine does not guarantee that. In summary,
\begin{enumerate}
\item A \ac{dnp} is a conditional distribution over functions trained to model the empirical conditional distributions of functions $f \sim P$.
\item A \ac{dnp} is permutation invariant in $O$ and $T$.
\item A CNP is scalable, achieving a running time complexity of $\mathcal{O}(n+m)$ for making $m$ predictions with $n$ observations.
\end{enumerate}

Within this specification of the model there are still some aspects that can be modified to suit specific requirements. The exact implementation of $h$, for example, can be adapted to the data type. For low dimensional data the encoder can be implemented as an MLP, whereas for inputs with larger dimensions and spatial correlations it can also include convolutions. 
Finally, in the setup described the model is not able to produce any coherent samples, as it learns to model only a factored prediction of the mean and the variances, disregarding the covariance between target points. This is a result of this particular implementation of the model. One way we can obtain coherent samples is by introducing a latent variable that we can sample from. We carry out some proof-of-concept experiments on such a model in section~\ref{latent_var_model}.

\section{Related research}
\label{sec:related}

\subsection{Gaussian Processes}
The goal of our research is to incorporate ideas from GP inference into a NN training regime to overcome certain drawbacks of both. There are a number of papers that address some of the same issues within the GP framework. Scaling issues with GPs have been addressed by sparse GPs~\citep{snelson2006sparse}, while overcoming the limited expressivity resulting from functional restrictions is the motivation for Deep GPs~\citep{damianou2013deep, salimbeni2017doubly}. The authors of Deep Kernel learning~\citep{wilson2016deep}, also combine ideas from DL and GPs. Their model, however, remains closer to GPs as the neural network is used to learn more expressive kernels that are then used within a GP.

\subsection{Meta-Learning}
Deep learning models are generally more scalable and are very successful at learning features and prior knowledge from the data directly. However they tend to be less flexible with regards to input size and order. Additionally, in general they only approximate one function as opposed to distributions over functions. Meta-learning approaches address the latter and share our core motivations. Recently meta-learning has been applied to a wide range of tasks like RL~\citep{wang2016learning, finn2017model} or program induction~\citep{devlin2017neural}.

Often meta-learning algorithms are implemented as deep generative models that learn to do few-shot estimations of the underlying density of the data. Generative Query Networks (GQN), for example, predict new viewpoints in 3D scenes given some context observations using a similar training regime to NPs~\citep{eslami2018neural}. As such, NPs can be seen as a generalisation of GQN to few-shot prediction tasks beyond scene understanding, such as regression and classification. 
Another way of carrying out few-shot density estimation is by updating existing models like PixelCNN~\citep{van2016conditional} and augmenting them with attention mechanisms~\citep{reed2017few} or including a memory unit in a VAE model~\citep{bornschein2017variational}. Another successful latent variable approach is to explicitly condition on some context during inference~\citep{rezende2016one}. Given the generative nature of these models they are usually applied to image generation tasks, but models that include a conditioning class-variable can be used for classification as well.
    
Classification itself is another common task in meta-learning. Few-shot classification algorithms usually rely on some distance metric in feature space to compare target images to the observations provided ~\citep{koch2015siamese}, ~\citep{santoro2016one}. Matching networks~\citep{vinyals2016matching, bartunov2016fast} are closely related to CNPs. In their case features of samples are compared with target features using an attention kernel. At a higher level one can interpret this model as a CNP where the aggregator is just the concatenation over all input samples and the decoder $g$ contains an explicitly defined distance kernel. In this sense matching networks are closer to GPs than to CNPs, since they require the specification of a distance kernel that CNPs learn from the data instead. In addition, as MNs carry out all-to-all comparisons they scale with $\mathcal{O}(n \times m)$, although they can be modified to have the same complexity of $\mathcal{O}(n + m)$ as CNPs~\citep{snell2017prototypical}. 

A model that is conceptually very similar to CNPs (and in particular the latent variable version) is the ``neural statistician'' paper~\citep{edwards2016towards} and the related variational homoencoder~\citep{hewitt2018variational}. As with the other generative models the neural statistician learns to estimate the density of the observed data but does not allow for targeted sampling at what we have been referring to as input positions $x_i$. Instead, one can only generate i.i.d. samples from the estimated density. Finally, the latent variant of CNP can also be seen as an approximated amortized version of Bayesian DL \cite{gal2016dropout, blundell2015weight, louizos2017bayesian, louizos2017multiplicative}

\section{Experimental Results}
\label{sec:experiments}

\begin{figure}[ht]
\vskip 0.2in
\begin{center}
\centerline{\includegraphics[width=\columnwidth]{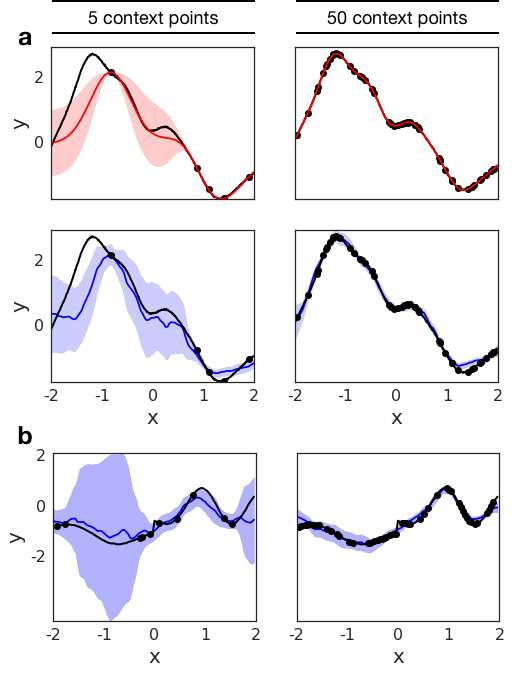}}
\caption{\textbf{1-D Regression}. Regression results on a 1-D curve (black line) using 5 (left column) and 50 (right column) context points (black dots). The first two rows show the predicted mean and variance for the regression of a single underlying kernel for GPs (red) and CNPs (blue). The bottom row shows the predictions of CNPs for a curve with switching kernel parameters.}
\label{regression_results}
\end{center}
\vskip -0.2in
\end{figure}

\begin{figure*}[ht]
\vskip 0.2in
\begin{center}
\centerline{\includegraphics[width=\columnwidth*2]{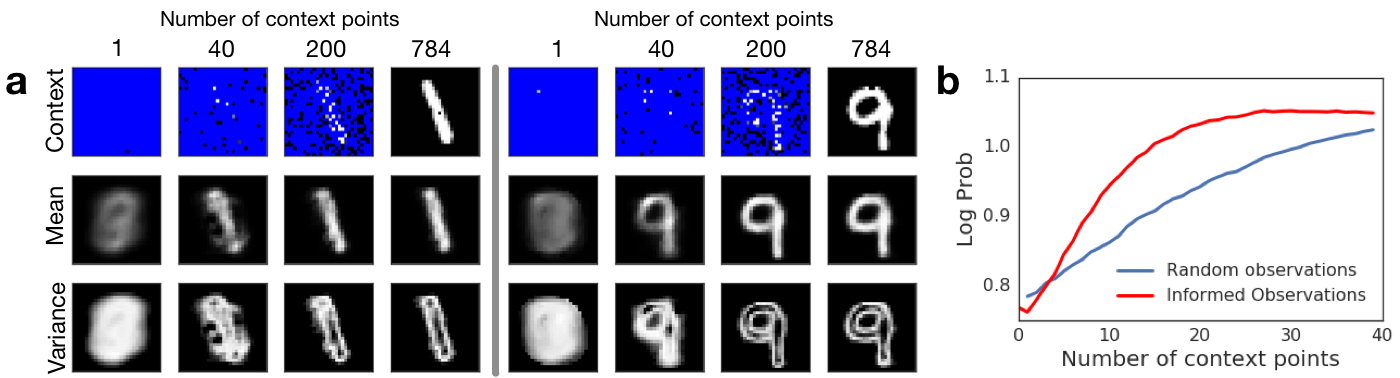}}
\caption{\textbf{Pixel-wise image regression on MNIST.} Left: Two examples of image regression with varying numbers of observations. We provide the model with 1, 40, 200 and 728 context points (top row) and query the entire image. The resulting mean (middle row) and variance (bottom row) at each pixel position is shown for each of the context images. Right: model accuracy with increasing number of observations that are either chosen at random (blue) or by selecting the pixel with the highest variance (red).}
\label{ic_mnist_results}
\end{center}
\vskip -0.2in
\end{figure*}

\subsection{Function Regression}
\label{sec:regression}

As a first experiment we test CNP on the classical 1D regression task that is used as a common baseline for GPs. We generate two different datasets that consist of functions generated from a GP with an exponential kernel. In the first dataset we use a kernel with fixed parameters, and in the second dataset the function switches at some random point on the real line between two functions each sampled with different kernel parameters.

At every training step we sample a curve from the GP, select a subset of $n$ points $(x_i,y_i)$ as observations, and a subset of points $(x_t, y_t)$ as target points.
Using the model described in Figure \ref{fig:model}, 
the observed points are encoded using a three layer MLP encoder $h$ with a $128$ dimensional output representation $r_i$. The representations are aggregated into a single representation $r = \frac{1}{n}\sum r_i$  which is concatenated to ${x_t}$ and passed to a decoder $g$ consisting of a five layer MLP. The decoder outputs a Gaussian mean and variance for the target outputs $\hat{y}_t$. 
We train the model to maximize the log-likelihood of the target points using the Adam optimizer~\cite{kingma2014adam}.


Two examples of the regression results obtained for each of the datasets are shown in Figure~\ref{regression_results}.
We compare the model to the predictions generated by a GP with the correct hyperparameters, which constitutes an upper bound on our performance. Although the prediction generated by the GP is smoother than the CNP's prediction both for the mean and variance, the model is able to learn to regress from a few context points for both the fixed kernels and switching kernels. As the number of context points grows, the accuracy of the model improves and the approximated uncertainty of the model decreases. Crucially, we see the model learns to estimate its own uncertainty given the observations very accurately.  Nonetheless it provides a good approximation that increases in accuracy as the number of context points increases.

Furthermore the model achieves similarly good performance on the switching kernel task. This type of regression task is not trivial for GPs whereas in our case we only have to change the dataset used for training. 


\subsection{Image Completion}
\label{sec:completion}

We consider image completion as a regression task over functions in either $f: [0,1]^2 \to [0,1]$ for grayscale images, or $f: [0,1]^2 \to [0,1]^3$ for RGB images. The input $x$ is the 2D pixel coordinates normalized to $[0,1]^2$, and the output $y$ is either the grayscale intensity or a vector of the RGB intensities of the corresponding pixel.
For this completion task we use exactly the same model architecture as for 1D function regression (with the exception of making the last layer 3-dimensional for RGB).

We test CNP on two different data sets: the MNIST handwritten digit database~\cite{lecun1998gradient} and large-scale CelebFaces Attributes (CelebA) dataset~\cite{liu2015faceattributes}. The model and training procedure are the same for both:
at each step we select an image from the dataset and pick a subset of the pixels as observations. Conditioned on these, the model is trained to predict the values of all the pixels in the image (including the ones it has been conditioned on). Like in 1D regression, the model outputs a Gaussian mean and variance for each pixel and is optimized with respect to the log-likelihood of the ground-truth image.
It is important to point out that we choose images as our dataset because they constitute a complex 2-D function that is easy to evaluate visually, not to compare to generative models benchmarks.

\subsubsection{MNIST}
We first test CNP on the MNIST dataset and use the test set to evaluate its performance. As shown in Figure~\ref{ic_mnist_results}a the model learns to make good predictions of the underlying digit even for a small number of context points. Crucially, when conditioned only on one non-informative context point (e.g. a black pixel on the edge) the model's prediction corresponds to the average over all MNIST digits. As the number of context points increases the predictions become more similar to the underlying ground truth. This demonstrates the model's capacity to extract dataset specific prior knowledge.
It is worth mentioning that even with a complete set of observations the model does not achieve pixel-perfect reconstruction, as we have a bottleneck at the representation level. 

Since this implementation of CNP returns factored outputs, the best prediction it can produce given limited context information is to average over all possible predictions that agree with the context. An alternative to this is to add latent variables in the model such that they can be sampled conditioned on the context to produce predictions with high probability in the data distribution. We consider this model later in section~\ref{latent_var_model}.

An important aspect of the model is its ability to estimate the uncertainty of the prediction. As shown in the bottom row of Figure~\ref{ic_mnist_results}a, as we add more observations, the variance shifts from being almost uniformly spread over the digit positions to being localized around areas that are specific to the underlying digit, specifically its edges. Being able to model the uncertainty given some context can be helpful for many tasks. One example is active exploration, where the model has a choice over where to observe. We test this by comparing the predictions of CNP when the observations are chosen according to  uncertainty (i.e.\ the pixel with the highest variance at each step), versus random pixels (Figure~\ref{ic_mnist_results}b). This method is a very simple way of doing active exploration, but it already produces better prediction results than selecting the conditioning points at random.

\begin{figure}[ht]
\vskip 0.2in
\begin{center}
\centerline{\includegraphics[width=0.85\columnwidth]{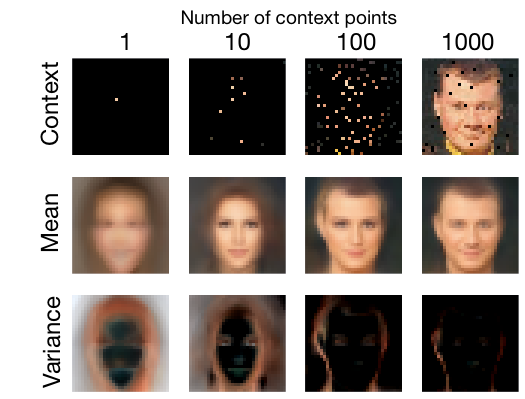}}
\caption{\textbf{Pixel-wise image completion on CelebA.} Two examples of CelebA image regression with varying numbers of observations. We provide the model with 1, 10, 100 and 1000 context points (top row) and query the entire image. The resulting mean (middle row) and variance (bottom row) at each pixel position is shown for each of the context images. }
\label{ic_celeba_results}
\end{center}
\vskip -0.2in
\end{figure}

\begin{figure}[ht]
\vskip 0.2in
\begin{center}
\centerline{\includegraphics[width=0.8\columnwidth]{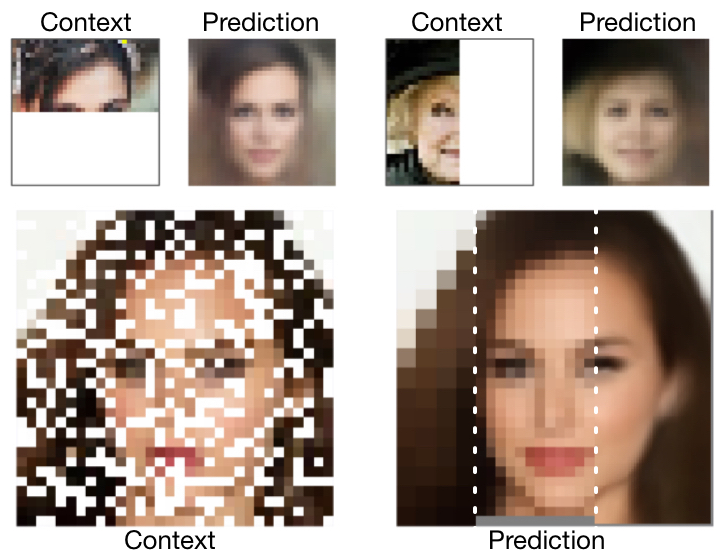}}
\caption{\textbf{Flexible image completion}. In contrast to standard conditional models, CNPs can be directly conditioned on observed pixels in arbitrary patterns, even ones which were never seen in the training set. Similarly, the model can predict values for pixel coordinates that were never included in the training set, like subpixel values in different resolutions. The dotted white lines were added for clarity after generation.}
\label{flexible_celeba}
\end{center}
\vskip -0.2in
\end{figure}

\begin{figure}[ht]
\vskip 0.2in
\begin{center}
\centerline{\includegraphics[width=0.5\columnwidth]{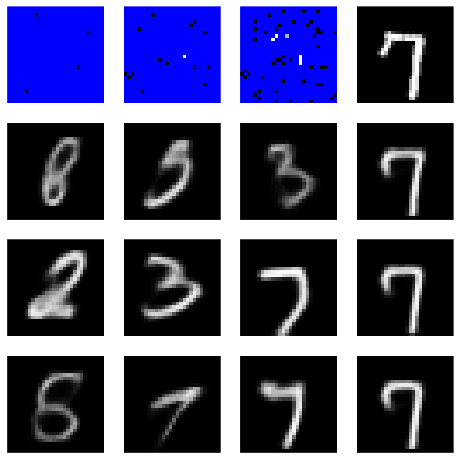} \includegraphics[width=0.5\columnwidth]{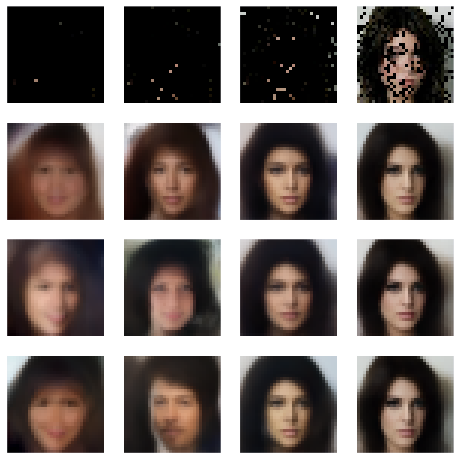}}
\caption{\textbf{Image completion with a latent variable model}. The latent variables capture the global uncertainty, allowing the sampling of different coherent images which conform to the observations. As the number of observations increases, uncertainty is reduced and the samples converge to a single estimate.}
\label{latent}
\end{center}
\vskip -0.2in
\end{figure}

\subsubsection{CelebA}
We also apply CNP to CelebA, a dataset of images of celebrity faces, and report performance obtained on the test set.
As shown in Figure~\ref{ic_celeba_results} our model is able to capture the complex shapes and colours of this dataset with predictions conditioned on less than 10\% of the pixels being already close to ground truth. As before, given few context points the model averages over all possible faces, but as the number of context pairs increases the predictions capture image-specific details like face orientation and facial expression. Furthermore, as the number of context points increases the variance is shifted towards the edges in the image.

An important aspect of CNPs demonstrated in Figure~\ref{flexible_celeba}, is its flexibility not only in the number of observations and targets it receives but also with regards to their input values. It is interesting to compare this property to GPs on one hand, and to trained generative models \cite{van2016conditional, gregor2015draw} on the other hand.

The first type of flexibility can be seen when conditioning on subsets that the model has not encountered during training. Consider conditioning the model on one half of the image, fox example. This forces the model to not only predict pixel values according to some stationary smoothness property of the images, but also according to global spatial properties, e.g. symmetry and the relative location of different parts of faces. As seen in the first row of the figure, CNPs are able to capture those properties. A GP with a stationary kernel cannot capture this, and in the absence of observations would revert to its mean (the mean itself can be non-stationary but usually this would not be enough to capture the interesting properties).

In addition, the model is flexible with regards to the target input values. This means, e.g., we can query the model at resolutions it has not seen during training. We take a model that has only been trained using pixel coordinates of a specific resolution, and predict at test time subpixel values for targets between the original coordinates. As shown in Figure~\ref{flexible_celeba}, with one forward pass we can query the model at different resolutions. While GPs also exhibit this type of flexibility, it is  not the case for trained generative models, which can only predict values for the pixel coordinates on which they were trained.
In this sense, CNPs capture the best of both worlds -- it is flexible in regards to the conditioning and prediction task, and has the capacity to extract domain knowledge from a training set.

We compare CNPs quantitatively to two related models: kNNs and GPs. As shown in Table~\ref{image_recon} CNPs outperform the latter when number of context points is small (empirically when half of the image or less is provided as context). When the majority of the image is given as context exact methods like GPs and kNN will perform better. From the table we can also see that the order in which the context points are provided is less important for CNPs, since providing the context points in order from top to bottom still results in good performance. Both insights point to the fact that CNPs learn a data-specific `prior' that will generate good samples even when the number of context points is very small.

\subsubsection{Latent variable model}
\label{latent_var_model}

The main model we use throughout this paper is a factored model that predicts the mean and variance of the target outputs. Although we have shown that the mean is by itself a useful prediction, and that the variance is a good way to capture the uncertainty, this factored model prevents us from obtaining coherent samples over multiple targets.

Consider the MNIST experiment in Figure~\ref{ic_mnist_results}, conditioned on a small number of observations. Rather than predicting only the mean of all digits, sometimes we need a model that can be used to sample different coherent images of all the possible digits conditioned on the observations. 

GPs can do this because they contain a parametric kernel predicting the covariance between all the points in the observations and targets. This forms a multivariate Gaussian which can be used to coherently draw samples. 
In order to maintain this property in a trained model, one approach is to train the model to predict a GP kernel \cite{wilson2016deep}. However the difficulty is the need to back-propagate through the sampling which involves a large matrix inversion (or some approximation of it).

\begin{table}[H]
\begin{center}
 \begin{tabular}{c | c c c | c c c} 
 \toprule
  & \multicolumn{3}{c}{Random Context} & \multicolumn{3}{c}{Ordered Context} \\ [0.5ex] 
  \# & 10 & 100 & 1000 & 10 & 100 & 1000\\
 \midrule
  \small{kNN} & 0.215 & 0.052 & 0.007 & 0.370 & 0.273 & 0.007\\ 
  \small{GP}  & 0.247 & 0.137 & \textbf{0.001} & 0.257 & 0.220 & \textbf{0.002}\\
  \small{\ac{dnp}}  & \textbf{0.039} & \textbf{0.016} & 0.009 & \textbf{0.057} & \textbf{0.047} & 0.021\\
 \bottomrule
\end{tabular}
\label{image_recon}
\caption{Pixel-wise mean squared error for all of the pixels in the image completion task on the CelebA data set with increasing number of context points (10, 100, 1000). The context points are chosen either at random or ordered from the top-left corner to the bottom-right. With fewer context points CNPs outperform kNNs and GPs. In addition CNPs perform well regardless of the order of the context points, whereas GPs and kNNs perform worse when the context is ordered.}
\end{center}
\end{table}

In contrast, the approach we use is to simply add latent variables $z$ to our decoder $g$, allowing our model to capture global uncertainty. In order to generate a coherent sample, we compute the representation $r$ from the observations, which parametrizes a Gaussian distribution over the latents $z$. $z$ is then sampled once and used to generate the predictions for all targets. To get a different coherent sample we draw a new sample from the latents $z$ and run the decoder again for all targets.
Similar to the standard VAE \cite{kingma2013auto, rezende2014stochastic} we train this model by optimizing a variational lower bound of the log-likelihood, using a conditional Gaussian prior $p(z|O)$ that is conditioned on the observations, and a Gaussian posterior $p(z|O, T)$ that is also conditioned on the target points.

We apply this model to MNIST and CelebA (Figure~\ref{latent}). We use the same models as before, but we concatenate the representation $r$ to a vector of latent variables $z$ of size $64$ (for CelebA we use bigger models where the sizes of $r$ and $z$ are $1024$ and $128$ respectively).
For both the prior and posterior models, we use three layered MLPs and average their outputs. We emphasize that the difference between the prior and posterior is that the prior only sees the observed pixels, while the posterior sees both the observed and the target pixels.
When sampling from this model with a small number of observed pixels, we get coherent samples and we see that the variability of the datasets is captured. As the model is conditioned on more and more observations, the variability of the samples drops and they eventually converge to a single possibility.

\subsection{Classification}
\label{sec:classification}
\begin{figure}[ht]
\vskip 0.2in
\begin{center}
\centerline{\includegraphics[width=\columnwidth]{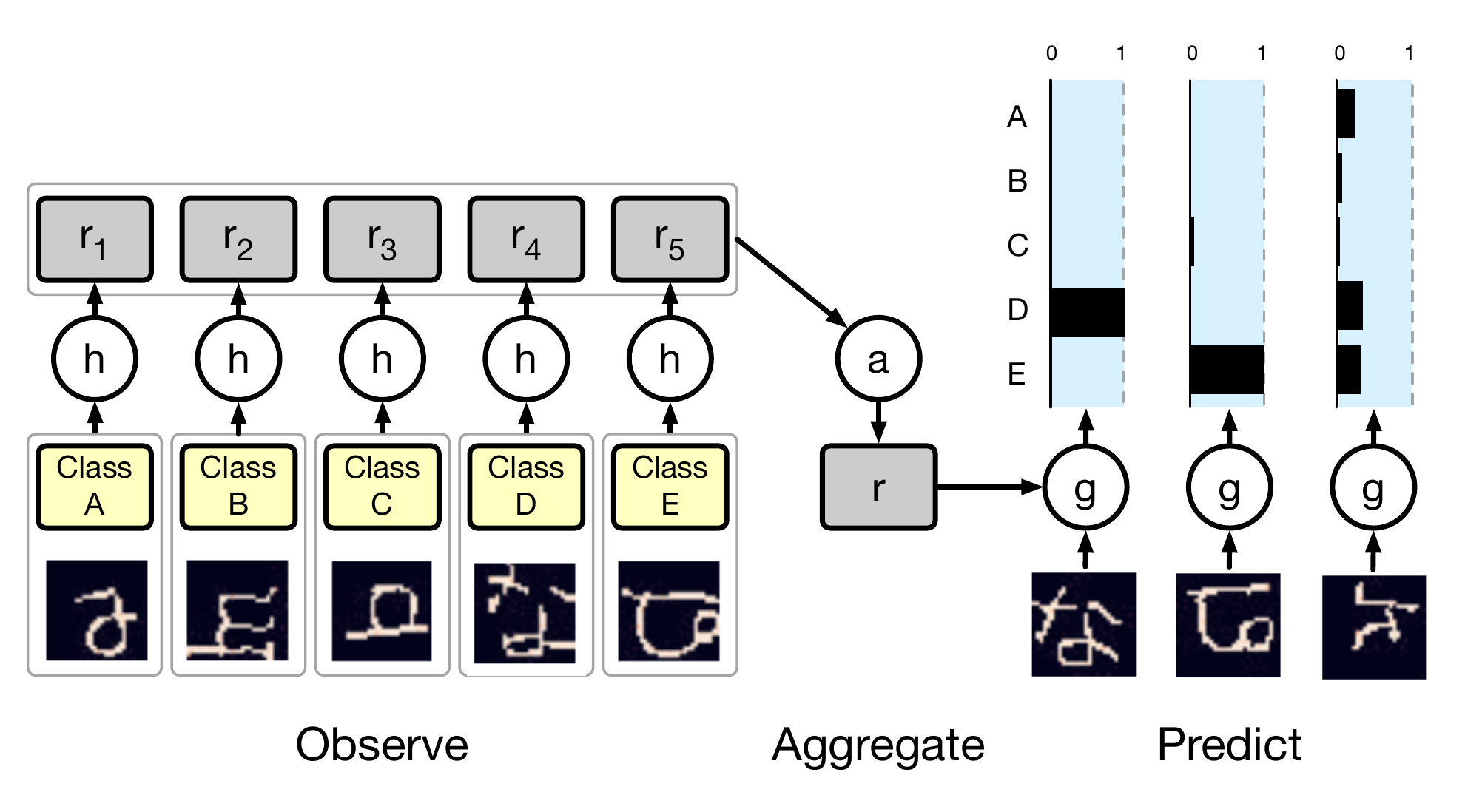}}
\caption{\textbf{One-shot Omniglot classification}. At test time the model is presented with a labelled example for each class, and outputs the classification probabilities of a new unlabelled example. 
The model uncertainty grows when the new example comes from an un-observed class.}

\label{classification}
\end{center}
\vskip -0.2in
\end{figure}

Finally, we apply the model to one-shot classification using the Omniglot dataset~\cite{lake2015human} (see Figure~\ref{classification} for an overview of the task). This dataset consists of 1,623 classes of characters from 50 different alphabets. Each class has only 20 examples and as such this dataset is particularly suitable for few-shot learning algorithms. As in~\citep{vinyals2016matching} we use 1,200 randomly selected classes as our training set and the remainder as our testing data set. In addition we augment the dataset following the protocol described in~\cite{santoro2016one}.
This includes cropping the image from $32\times32$ to $28\times28$, applying small random translations and rotations to the inputs, and also increasing the number of classes by rotating every character by 90 degrees and defining that to be a new class. We generate the labels for an N-way classification task by choosing N random classes at each training step and arbitrarily assigning the labels $0, ..., N-1$ to each.

Given that the input points are images, we modify the architecture of the encoder $h$ to include convolution layers as mentioned in section~\ref{sec:model}. In addition we only aggregate over inputs of the same class by using the information provided by the input label. The aggregated class-specific representations are then concatenated to form the final representation. Given that both the size of the class-specific representations and the number of classes are constant, the size of the final representation is still constant and thus the $\mathcal{O}(n+m)$ runtime still holds.

The results of the classification are summarized in Table~\ref{table_results}. CNPs achieve higher accuracy than models that are significantly more complex (like MANN). While CNPs do not beat state of the art for one-shot classification our accuracy values are comparable. Crucially, we reach those values using a significantly simpler architecture (three convolutional layers for the encoder and a three-layer MLP for the decoder) and with a lower runtime of $\mathcal{O}(n+m)$ at test time as opposed to $\mathcal{O}(nm)$.  
\begin{table}
\begin{center}
 \begin{tabular}{l | c c c c l} 
 \toprule
  & \multicolumn{2}{c}{5-way Acc} & \multicolumn{2}{c}{20-way Acc} &\small{Runtime} \\ [0.5ex] 
  & 1-shot & 5-shot & 1-shot & 5-shot &\\
 \midrule
  \small{MANN} & 82.8\% & 94.9\% & -  & - &\tiny{$\mathcal{O}(nm)$}\\ 
  \small{MN} & \textbf{98.1\%} & \textbf{98.9\%} & \textbf{93.8\%}  & \textbf{98.5\%} &\tiny{$\mathcal{O}(nm)$}\\
  \small{\ac{dnp}} & 95.3\% & 98.5\% & 89.9\%  & 96.8\% &\tiny{$\mathcal{O}(n+m)$}\\
 \bottomrule
\end{tabular}

\end{center}
\caption{\textbf{Classification results on Omniglot}. Results on the same task for MANN~\citep{santoro2016one}, and matching networks (MN)~\citep{vinyals2016matching} and CNP.}
\label{table_results}
\end{table}

\section{Discussion}
\label{sec:discussion}
In this paper we have introduced \aclp{dnp}, a model that is both flexible at test time and has the capacity to extract prior knowledge from training data.
We have demonstrated its ability to perform a variety of tasks including regression, classification and image completion.
We compared CNPs to Gaussian Processes on one hand, and deep learning methods on the other, and also discussed the relation to meta-learning and few-shot learning.

It is important to note that the specific \ac{dnp} implementations described here are just simple proofs-of-concept and can be substantially extended, e.g.\ by including more elaborate architectures in line with modern deep learning advances.

To summarize, this work can be seen as a step towards learning high-level abstractions, one of the grand challenges of contemporary machine learning. Functions learned by most conventional deep learning models are tied to a specific, constrained statistical context at any stage of training. A trained CNP is more general, in that it encapsulates the high-level statistics of a family of functions. As such it constitutes a high-level abstraction that can be reused for multiple tasks. In future work we will explore how far these models can help in tackling the many key machine learning problems that seem to hinge on abstraction, such as transfer learning, meta-learning, and data efficiency.

\section*{Acknowledgements}
We would like to thank Shakir Mohamed, Fabio Viola, Oriol Vinyals, Irene Garnelo, Daniel Burgess, Kevin McKee and Ellen Clancy for insightful discussions and being pretty cool.

\bibliography{references}

\begin{thebibliography}{33}
\providecommand{\natexlab}[1]{#1}
\providecommand{\url}[1]{\texttt{#1}}
\expandafter\ifx\csname urlstyle\endcsname\relax
  \providecommand{\doi}[1]{doi: #1}\else
  \providecommand{\doi}{doi: \begingroup \urlstyle{rm}\Url}\fi

\bibitem[Bartunov \& Vetrov(2016)Bartunov and Vetrov]{bartunov2016fast}
Bartunov, S. and Vetrov, D.~P.
\newblock Fast adaptation in generative models with generative matching
  networks.
\newblock \emph{arXiv preprint arXiv:1612.02192}, 2016.

\bibitem[Blundell et~al.(2015)Blundell, Cornebise, Kavukcuoglu, and
  Wierstra]{blundell2015weight}
Blundell, C., Cornebise, J., Kavukcuoglu, K., and Wierstra, D.
\newblock Weight uncertainty in neural networks.
\newblock \emph{arXiv preprint arXiv:1505.05424}, 2015.

\bibitem[Bornschein et~al.(2017)Bornschein, Mnih, Zoran, and
  J.~Rezende]{bornschein2017variational}
Bornschein, J., Mnih, A., Zoran, D., and J.~Rezende, D.
\newblock Variational memory addressing in generative models.
\newblock In \emph{Advances in Neural Information Processing Systems}, pp.\
  3923--3932, 2017.

\bibitem[Damianou \& Lawrence(2013)Damianou and Lawrence]{damianou2013deep}
Damianou, A. and Lawrence, N.
\newblock Deep gaussian processes.
\newblock In \emph{Artificial Intelligence and Statistics}, pp.\  207--215,
  2013.

\bibitem[Devlin et~al.(2017)Devlin, Bunel, Singh, Hausknecht, and
  Kohli]{devlin2017neural}
Devlin, J., Bunel, R.~R., Singh, R., Hausknecht, M., and Kohli, P.
\newblock Neural program meta-induction.
\newblock In \emph{Advances in Neural Information Processing Systems}, pp.\
  2077--2085, 2017.

\bibitem[Edwards \& Storkey(2016)Edwards and Storkey]{edwards2016towards}
Edwards, H. and Storkey, A.
\newblock Towards a neural statistician.
\newblock 2016.

\bibitem[Eslami et~al.(2018)Eslami, Rezende, Besse, Viola, Morcos, Garnelo,
  Ruderman, Rusu, Danihelka, Gregor, et~al.]{eslami2018neural}
Eslami, S.~A., Rezende, D.~J., Besse, F., Viola, F., Morcos, A.~S., Garnelo,
  M., Ruderman, A., Rusu, A.~A., Danihelka, I., Gregor, K., et~al.
\newblock Neural scene representation and rendering.
\newblock \emph{Science}, 360\penalty0 (6394):\penalty0 1204--1210, 2018.

\bibitem[Finn et~al.(2017)Finn, Abbeel, and Levine]{finn2017model}
Finn, C., Abbeel, P., and Levine, S.
\newblock Model-agnostic meta-learning for fast adaptation of deep networks.
\newblock \emph{arXiv preprint arXiv:1703.03400}, 2017.

\bibitem[Gal \& Ghahramani(2016)Gal and Ghahramani]{gal2016dropout}
Gal, Y. and Ghahramani, Z.
\newblock Dropout as a bayesian approximation: Representing model uncertainty
  in deep learning.
\newblock In \emph{international conference on machine learning}, pp.\
  1050--1059, 2016.

\bibitem[Garnelo et~al.(2016)Garnelo, Arulkumaran, and
  Shanahan]{garnelo2016towards}
Garnelo, M., Arulkumaran, K., and Shanahan, M.
\newblock Towards deep symbolic reinforcement learning.
\newblock \emph{arXiv preprint arXiv:1609.05518}, 2016.

\bibitem[Gregor et~al.(2015)Gregor, Danihelka, Graves, Rezende, and
  Wierstra]{gregor2015draw}
Gregor, K., Danihelka, I., Graves, A., Rezende, D.~J., and Wierstra, D.
\newblock Draw: A recurrent neural network for image generation.
\newblock \emph{arXiv preprint arXiv:1502.04623}, 2015.

\bibitem[Hewitt et~al.(2018)Hewitt, Gane, Jaakkola, and
  Tenenbaum]{hewitt2018variational}
Hewitt, L., Gane, A., Jaakkola, T., and Tenenbaum, J.~B.
\newblock The variational homoencoder: Learning to infer high-capacity
  generative models from few examples.
\newblock 2018.

\bibitem[J.~Rezende et~al.(2016)J.~Rezende, Danihelka, Gregor, Wierstra,
  et~al.]{rezende2016one}
J.~Rezende, D., Danihelka, I., Gregor, K., Wierstra, D., et~al.
\newblock One-shot generalization in deep generative models.
\newblock In \emph{International Conference on Machine Learning}, pp.\
  1521--1529, 2016.

\bibitem[Kingma \& Ba(2014)Kingma and Ba]{kingma2014adam}
Kingma, D.~P. and Ba, J.
\newblock Adam: A method for stochastic optimization.
\newblock \emph{arXiv preprint arXiv:1412.6980}, 2014.

\bibitem[Kingma \& Welling(2013)Kingma and Welling]{kingma2013auto}
Kingma, D.~P. and Welling, M.
\newblock Auto-encoding variational bayes.
\newblock \emph{arXiv preprint arXiv:1312.6114}, 2013.

\bibitem[Koch et~al.(2015)Koch, Zemel, and Salakhutdinov]{koch2015siamese}
Koch, G., Zemel, R., and Salakhutdinov, R.
\newblock Siamese neural networks for one-shot image recognition.
\newblock In \emph{ICML Deep Learning Workshop}, volume~2, 2015.

\bibitem[Lake et~al.(2015)Lake, Salakhutdinov, and Tenenbaum]{lake2015human}
Lake, B.~M., Salakhutdinov, R., and Tenenbaum, J.~B.
\newblock Human-level concept learning through probabilistic program induction.
\newblock \emph{Science}, 350\penalty0 (6266):\penalty0 1332--1338, 2015.

\bibitem[Lake et~al.(2017)Lake, Ullman, Tenenbaum, and
  Gershman]{lake2017building}
Lake, B.~M., Ullman, T.~D., Tenenbaum, J.~B., and Gershman, S.~J.
\newblock Building machines that learn and think like people.
\newblock \emph{Behavioral and Brain Sciences}, 40, 2017.

\bibitem[LeCun et~al.(1998)LeCun, Bottou, Bengio, and
  Haffner]{lecun1998gradient}
LeCun, Y., Bottou, L., Bengio, Y., and Haffner, P.
\newblock Gradient-based learning applied to document recognition.
\newblock \emph{Proceedings of the IEEE}, 86\penalty0 (11):\penalty0
  2278--2324, 1998.

\bibitem[Liu et~al.(2015)Liu, Luo, Wang, and Tang]{liu2015faceattributes}
Liu, Z., Luo, P., Wang, X., and Tang, X.
\newblock Deep learning face attributes in the wild.
\newblock In \emph{Proceedings of International Conference on Computer Vision
  (ICCV)}, December 2015.

\bibitem[Louizos \& Welling(2017)Louizos and
  Welling]{louizos2017multiplicative}
Louizos, C. and Welling, M.
\newblock Multiplicative normalizing flows for variational bayesian neural
  networks.
\newblock \emph{arXiv preprint arXiv:1703.01961}, 2017.

\bibitem[Louizos et~al.(2017)Louizos, Ullrich, and
  Welling]{louizos2017bayesian}
Louizos, C., Ullrich, K., and Welling, M.
\newblock Bayesian compression for deep learning.
\newblock In \emph{Advances in Neural Information Processing Systems}, pp.\
  3290--3300, 2017.

\bibitem[Rasmussen \& Williams(2004)Rasmussen and
  Williams]{rasmussen2004gaussian}
Rasmussen, C.~E. and Williams, C.~K.
\newblock Gaussian processes in machine learning.
\newblock In \emph{Advanced lectures on machine learning}, pp.\  63--71.
  Springer, 2004.

\bibitem[Reed et~al.(2017)Reed, Chen, Paine, Oord, Eslami, J.~Rezende, Vinyals,
  and de~Freitas]{reed2017few}
Reed, S., Chen, Y., Paine, T., Oord, A. v.~d., Eslami, S., J.~Rezende, D.,
  Vinyals, O., and de~Freitas, N.
\newblock Few-shot autoregressive density estimation: Towards learning to learn
  distributions.
\newblock 2017.

\bibitem[Rezende et~al.(2014)Rezende, Mohamed, and
  Wierstra]{rezende2014stochastic}
Rezende, D.~J., Mohamed, S., and Wierstra, D.
\newblock Stochastic backpropagation and approximate inference in deep
  generative models.
\newblock \emph{arXiv preprint arXiv:1401.4082}, 2014.

\bibitem[Salimbeni \& Deisenroth(2017)Salimbeni and
  Deisenroth]{salimbeni2017doubly}
Salimbeni, H. and Deisenroth, M.
\newblock Doubly stochastic variational inference for deep gaussian processes.
\newblock In \emph{Advances in Neural Information Processing Systems}, pp.\
  4591--4602, 2017.

\bibitem[Santoro et~al.(2016)Santoro, Bartunov, Botvinick, Wierstra, and
  Lillicrap]{santoro2016one}
Santoro, A., Bartunov, S., Botvinick, M., Wierstra, D., and Lillicrap, T.
\newblock One-shot learning with memory-augmented neural networks.
\newblock \emph{arXiv preprint arXiv:1605.06065}, 2016.

\bibitem[Snell et~al.(2017)Snell, Swersky, and Zemel]{snell2017prototypical}
Snell, J., Swersky, K., and Zemel, R.
\newblock Prototypical networks for few-shot learning.
\newblock In \emph{Advances in Neural Information Processing Systems}, pp.\
  4080--4090, 2017.

\bibitem[Snelson \& Ghahramani(2006)Snelson and Ghahramani]{snelson2006sparse}
Snelson, E. and Ghahramani, Z.
\newblock Sparse gaussian processes using pseudo-inputs.
\newblock In \emph{Advances in neural information processing systems}, pp.\
  1257--1264, 2006.

\bibitem[van~den Oord et~al.(2016)van~den Oord, Kalchbrenner, Espeholt,
  Vinyals, Graves, et~al.]{van2016conditional}
van~den Oord, A., Kalchbrenner, N., Espeholt, L., Vinyals, O., Graves, A.,
  et~al.
\newblock Conditional image generation with pixelcnn decoders.
\newblock In \emph{Advances in Neural Information Processing Systems}, pp.\
  4790--4798, 2016.

\bibitem[Vinyals et~al.(2016)Vinyals, Blundell, Lillicrap, Wierstra,
  et~al.]{vinyals2016matching}
Vinyals, O., Blundell, C., Lillicrap, T., Wierstra, D., et~al.
\newblock Matching networks for one shot learning.
\newblock In \emph{Advances in Neural Information Processing Systems}, pp.\
  3630--3638, 2016.

\bibitem[Wang et~al.(2016)Wang, Kurth-Nelson, Tirumala, Soyer, Leibo, Munos,
  Blundell, Kumaran, and Botvinick]{wang2016learning}
Wang, J.~X., Kurth-Nelson, Z., Tirumala, D., Soyer, H., Leibo, J.~Z., Munos,
  R., Blundell, C., Kumaran, D., and Botvinick, M.
\newblock Learning to reinforcement learn.
\newblock \emph{arXiv preprint arXiv:1611.05763}, 2016.

\bibitem[Wilson et~al.(2016)Wilson, Hu, Salakhutdinov, and
  Xing]{wilson2016deep}
Wilson, A.~G., Hu, Z., Salakhutdinov, R., and Xing, E.~P.
\newblock Deep kernel learning.
\newblock In \emph{Artificial Intelligence and Statistics}, pp.\  370--378,
  2016.

\end{thebibliography}
\bibliographystyle{icml2018}

\end{document}